\documentclass[10pt,twocolumn,letterpaper]{article}

\usepackage{ijcb}
\usepackage{times}
\usepackage{epsfig}
\usepackage{graphicx}
\usepackage{amsmath}
\usepackage{amssymb}
\usepackage[ruled, vlined, linesnumbered]{algorithm2e}
\usepackage[hyphens]{url}
\usepackage{hyperref}
\hypersetup{colorlinks=true,breaklinks=true}
\usepackage{breqn}
\usepackage{threeparttable}
\usepackage{tabularx}
\usepackage{multirow}
\usepackage{subcaption}
\usepackage[font={footnotesize},skip=4pt]{caption}
\usepackage[norule]{footmisc}

\newcolumntype{Y}{>{\centering\arraybackslash}X}

\ijcbfinalcopy % *** Uncomment this line for the final submission

\ifijcbfinal\fi

\makeatletter
\def\ps@IEEEtitlepagestyle{
\def\@oddfoot{\mycopyrightnotice}
\def\@evenfoot{}
}
\def\mycopyrightnotice{
{\hfill \footnotesize 978-1-6654-3780-6/21/\$31.00 \copyright 2021 IEEE\hfill}
}
\makeatother

\usepackage{fancyhdr}
\fancypagestyle{title}{%
  \setlength{\headheight}{22pt}%
  \fancyhf{}% No header/footer
  \fancyhead[C]{\small\begin{tabular}{@{}l}\Large\color{red} To appear in International Conference On Pattern Recognition (ICPR), 2020\end{tabular}}
}%

\begin{document}

%%%%%%%%% TITLE
\title{FedFace: Collaborative Learning of Face Recognition Model}

\author{Divyansh Aggarwal, Jiayu Zhou, Anil K. Jain\\
Michigan State University\\
East Lansing, MI, USA\\
{\tt\small \{aggarw49, jiayuz, jain\}@cse.msu.edu}
% \and
% Jiayu Zhou\\
% Michigan State University\\
% East Lansing, MI, USA\\
% {\tt\small jiayuz@msu.edu}

% \and
% Anil K. Jain\\
% Michigan State University\\
% East Lansing, MI, USA\\
% {\tt\small jain@cse.msu.edu}
}

\maketitle
\thispagestyle{empty}

%%%%%%%%% ABSTRACT

% \twocolumn[{%
% \renewcommand\twocolumn[1][]{#1}%
% \maketitle
%  \thispagestyle{empty}
% \begin{center}
%     \includegraphics[width=\linewidth]{images/main_figure.jpg}
% \end{center}
%     \captionof{Figure 1: }{An overview of the federated setup for training face recognition model. We tackle the scenario where each client is a mobile device with face images pertaining to only the user/owner of the device rather than data centers or organizations holding face datasets with multiple identities. The server sends the current global model to each of the participating client nodes who update the model based on their local training data. These local updates are transferred back to the server where they are aggregated to generate a global update. Through several rounds of communication between the servers and the client, a collaborative face recognition system can be obtained in a privacy preserving manner.}
%     \label{fig:Overview}
%     \vspace{1em}

% }]

\begin{abstract}
DNN-based face recognition models require large centrally aggregated face datasets for training. However, due to the growing data privacy concerns and legal restrictions, accessing and sharing face datasets has become exceedingly difficult. We propose FedFace, a federated learning (FL) framework for collaborative learning of face recognition models in a privacy aware manner. FedFace utilizes the face images available on multiple clients to learn an accurate and generalizable face recognition model where the face images stored at each client are neither shared with other clients nor the central host and each client is a mobile device containing face images pertaining to only the owner of the device (one identity per client). Our experiments show the effectiveness of FedFace in enhancing the verification performance of pre-trained face recognition system on standard face verification benchmarks namely LFW, IJB-A and IJB-C.
\end{abstract}

% \makeatletter
% \let\blfootnote\relax\footnotetext{\mycopyrightnotice}
% \makeatother
%%%%%%%%% BODY TEXT

\begin{figure*}[!t]
    \centering
    \captionsetup{font=footnotesize}
    \includegraphics[width=0.83\linewidth]{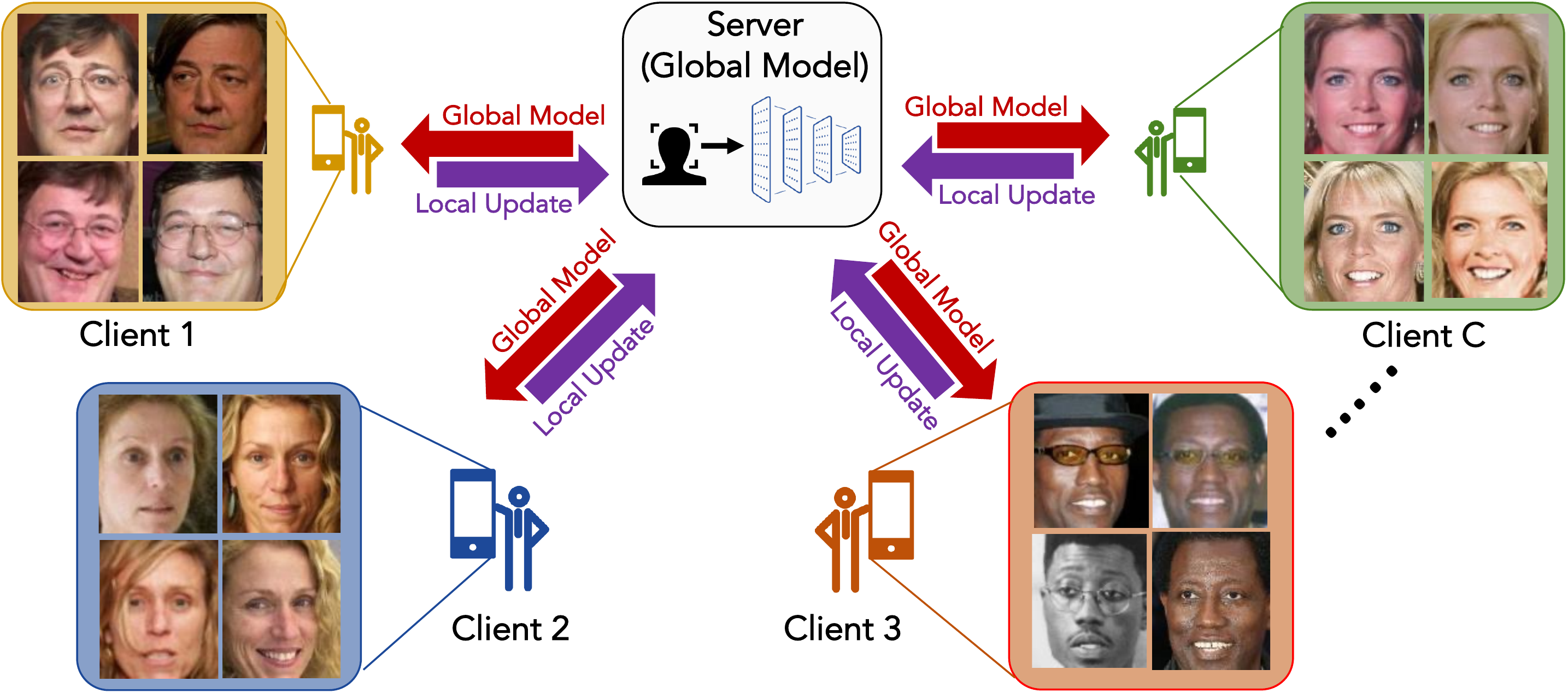}
    \caption{An overview of the federated setup for training face recognition models. We tackle the challenging but realistic scenario where each client is a mobile device with face images pertaining to only the user/owner of the device. In each communication round, the server sends the weights of the global model at that communication round to each of the participating client nodes who update the model based on their local training data. These local updates are transferred back to the server where they are aggregated to generate a global update. Through several rounds of communication between the server and the clients, a collaborative face recognition model can be learned in a privacy aware manner. Note that a client does not reveal its face images to the server or other clients.}
    \label{fig:overview}
\end{figure*}

\section{Introduction}

Automated Face Recognition (AFR) systems have been widely adopted for access control and person identification in a plethora of domains largely due to their accuracy, usability, and touchless and remote acquisition. AFR systems are embedded in our smartphones to unlock our phones, facilitate financial transactions as well as secure access to stored personal information. 
% They are also employed by law enforcement agencies for criminal identification and for recognizing and tracking criminals on a watchlist in surveillance footage from CCTV cameras~\cite{face_surveillance}. AFR systems are also deployed at airports to improve passenger throughput at various checkpoints~\cite{face_airports}. 
% In the United States, $26$ states allow law enforcement agencies to run searches against their databases of ID photos~\cite{US_face}. In India, Delhi police used face recognition system to trace $3000$ missing children in the span of $4$ days~\cite{Delhi_missing_children}. In Sydney, face recognition has been deployed at airports to help people move through security much faster and with ease. Apart from this, many retail stores in China and Japan are now using face recognition based payment solutions. This rapid growth of face recognition technology has motivated many computer vision researchers to build more accurate face recognition models.
According to the $2020$ NIST face recognition evaluation (FRVT)~\cite{nist_2020}, only $0.08$\% of searches in a database of $26.6$ million face images failed to find the true mate, compared to a failure rate of $4$\% in $2014$. That corresponds to a $50$-fold improvement in face recognition performance over the last six years. This significant boost in face recognition performance can be largely attributed to the use of Deep Neural Networks (DNNs) and large face training datasets for learning salient face representations.

DNN-based face recognition systems require large and diverse (in terms of race, gender and facial variations such as pose, illumination and expression) training face datasets to learn robust and generalizable face representations. However, due to the sensitive nature of biometric data (including face images), it is becoming exceedingly difficult for researchers to access large face datasets. According to a report by the United States Government Accountability Office (GAO), with the rapid proliferation of face images on social media websites such as Facebook, Twitter, and Instagram, face recognition models are being trained without obtaining the consent of the individuals whose face images are being used~\cite{gao_report}. A face recognition startup, Clearview AI, is currently facing litigation for allegedly amassing a dataset of about $3$ billion face images through internet scraping without acquiring consent from the individuals in the dataset~\cite{clearview}. Article $25$ of the General Data Protection Regulation (GDPR)~\cite{eu_gdpr}, enforces data protection ``by design and by default" in the development of any new framework and strictly prohibits collection and distribution of personal data without consent. Similar statutes exist in the states of California (California Consumer Privacy Act, CCPA)~\cite{ccpa} and Illinois (Personal Information Protection Act, PIPA)~\cite{pipa} to impose strict regulations on breach of personal data, including face images. Several large scale publicly available face datasets are now being retracted on account of containing face images of individuals without their approvals. Examples include Microsoft's MS Celeb~\cite{guo2016ms},  one of the largest face datasets with over $10M$ images from $10,000$ individuals which has been used by several commercial organizations (IBM, Alibaba, Nvidia and Sensetime) and academic research groups to train facial recognition systems. It was pulled off the internet in light of inclusion of face images of individuals without their consent~\cite{MSceleb_retracted}. Duke MTMC surveillance data set collected by Duke University researchers, and a Stanford University data set, called Brainwash, were also retracted from public domain on similar grounds~\cite{duke_mtmc, brainwash}. On account of growing and legitimate data privacy concerns on the ways training datasets for face recognition models can be collected and used, training AFR systems in a distributed and privacy aware manner is now becoming extremely important. Specifically, we need to directly learn salient features from face images on end user devices such as mobile phones while protecting their data (i.e, the face images on a client device are never revealed).

Federated learning (FL) provides a distributed and privacy-aware framework to train machine learning models where multiple client nodes collaboratively learn without sharing their data with the server or with other clients. McMahan~\etal~\cite{fedavg} proposed a method where the weights of the updated models generated by the participating clients after local training are communicated to the server; the server then aggregates these weights to obtain the weights for the global model. Variants of FedAvg~\cite{fedavg} such as FedProx~\cite{fedprox} and Agnostic Federated Learning~\cite{AFL} have also been proposed that aim at alleviating issues such as bias towards different clients when learning on clients with heterogeneous data. However, these conventional FL algorithms cannot be directly applied to an automated face recognition pipeline, since, the participating clients are mobile devices that only contain face images of the device owners, and thereby impose significant challenges in establishing discrimination among different faces.

In an effort to facilitate distributed training of face recognition systems from face images available on mobile devices, we propose a framework, called \emph{FedFace}, that learns an accurate face recognition model from multiple mobile devices in a collaborative manner without sending training face images outside of the device. Main contributions of our work are summarized as follows:

\begin{itemize}
    \item The \emph{FedFace} framework, for training face recognition systems in the federated setup to alleviate aggregating face datasets on a common server thereby facilitating data privacy. We tackle the challenging but realistic scenario where each of the participating clients has face images of only one identity.
    \item Our empirical results show that \emph{FedFace} enhanced the performance of a pre-trained face recognition system, CosFace~\cite{wang2018cosface}, by utilizing additional face data available on client nodes in a collaborative manner. \emph{FedFace} employs a mean feature initialization scheme for the class embeddings and a spreadout regularizer to ensure that the class embeddings are well separated.
    \item \emph{FedFace} is able to enhance the performance of a pre-trained face recognition system namely, CosFace, from a TAR of $81.43\%$ to $83.79\%$ on IJB-A @$0.1\%$ FAR and accuracy from $99.15\%$ to $99.28\%$ on LFW using the face images available on $1,000$ mobile devices in such a setup.
\end{itemize}
\begin{figure*}[!t]
    \centering
    \captionsetup{font=footnotesize}
    \includegraphics[width=0.9\linewidth]{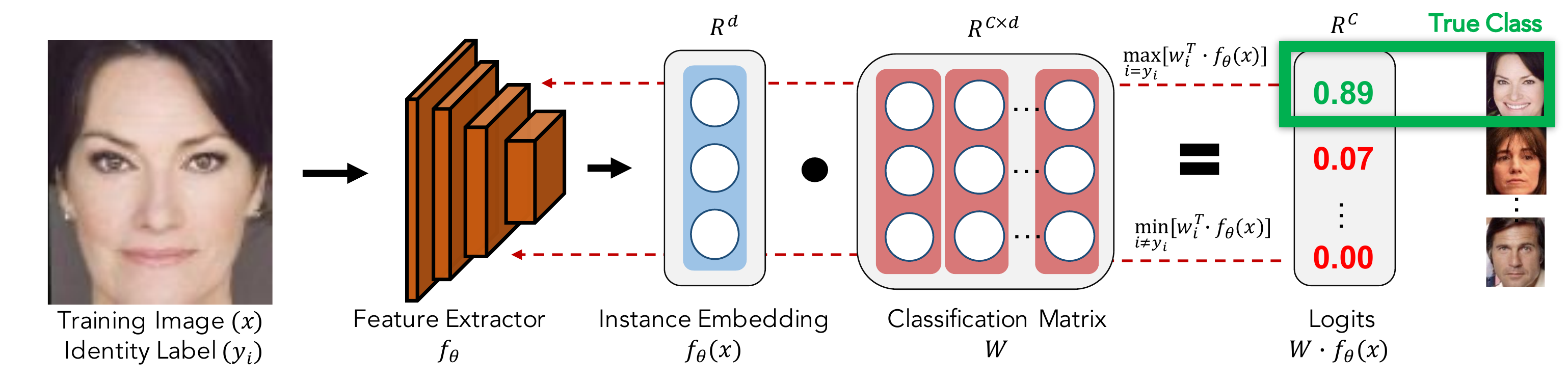}
    \caption{An overview of the training framework used by prevailing DNN-based AFR systems. An input $x$ with a label $y_i$ is passed through a feature extractor $f_{\theta}$ to obtain the feature vector $f_{\theta}(x)$. The feature vector is then multiplied with the classification matrix $W$ to get the logits or the likelihood of $x$ belonging to each of the $C$ identities. We then maximize the similarity between the feature vector and the positive class embedding $w_i$ and minimize the similarity between the feature vector and negative class embeddings (Red line indicates back-propagation of the loss through the model). In the FL setup, since each client does not have access to class embeddings of other clients/identities, the client cannot minimize the second term of the training objective. }
    \label{fig:problem_setup}
\end{figure*}
\section{Related Work}

Federated learning is a machine learning paradigm that enables training machine learning models collaboratively across multiple devices holding local data~\cite{fedavg,bonawitz2019towards,wei2020federated,augenstein2019generative,fedprox,AFL}. In the FL setup, the device that manages the overall model training is called the server and the devices that store the datasets and propagate their trained weights to the server are called client nodes. FL framework ensure that the local data of a client is never shared to other client nodes or the server. Therefore, by design, the FL framework keeps the data of the participating clients private and ensures security of the highly sensitive data. Several methods~\cite{kairouz2019advances,li2020federated} have been proposed in the literature to train deep learning models in the FL setting. Federated Averaging~\cite{fedavg} (FedAvg) proposed by McMahan \etal, which uses a weighted average of the local weight updates as the global weight update, is still one of the most widely adopted algorithms for training deep models in the FL setting.

FL has been widely adopted in a range of domains to train classification models in a privacy aware manner. Google uses FL in the Gboard mobile keyboard~\cite{hard2018federated,yang2018applied,chen2019federated,ramaswamy2019federated} for next word prediction and other natural language processing tasks. FL is also being used extensively in health informatics~\cite{huang2020loadaboost,sheller2020federated} as hospitals operate under strict privacy regulations~\cite{hippa} that prohibits their data from being shared with other organizations. However, its use in face recognition domain has been limited despite growing privacy concerns related to collection and sharing of face images. 

The only FL framework that we could find in the face domain is by Shao \etal~\cite{shao2020federated} who proposed a face presentation attack detection system, called FedPAD. They train a  generalizable face presentation attack detection system by learning from different spoof types and data distributions stored at different sites in the FL setup. \emph{However, to the best of our knowledge, no prior work has tackled the problem of learning a face recognition model under the FL setup.} More importantly, available FL algorithms cannot be directly applied to training face recognition models where the client nodes are mobile devices containing face images from only one identity. Yu~\etal~\cite{yu2020federated} proposed the Federated Averaging with Spreadout (FedAwS) algorithm that uses a geometric regularizer known as the Spreadout regularizer to tackle this problem for object recognition.

To the best of our knowledge, our work is the first to conduct a comprehensive evaluation of training face recognition models under the FL setup and to show that training with additional face images in a federated setting can help boost the performance of a pre-trained AFR system.

\section{FedFace}
\subsection{Problem Setup}
\label{section:problem_setup}
Face recognition systems are essentially embedding based classifiers \ie, both the classes (identities) and the input instance are embedded into a high dimensional Euclidean space. The similarity between the embedded input and the class embedding represents the likelihood of the instance belonging to that particular class. Formally, given an input face image $x$, the face feature extractor $f_{\theta}: X \rightarrow \mathcal{R}^d$ embeds the input instance into a $d$-dimensional feature vector $f_{\theta}(x)$. A fully connected layer with a softmax activation function parameterized by the class embedding matrix $W \in \mathcal{R}^{C\times d}$ ($C$ represents the total number of identities), then maps this feature vector to the logits or scores for each identity. The $k^{th}$ row of the class embedding matrix, $w_k$, is referred to as the class embedding of the $k^{th}$ class. The logit for the $k^{th}$ class is thus computed as $w_{k}^{T}\cdot f_{\theta}(x)$. A general overview of training a DNN-based AFR system is shown in Fig.~\ref{fig:problem_setup}. The training objective is as follows:

\vspace{-1\baselineskip}
\begin{multline}
    {l(x,y)} = \alpha \cdot (d(f_{\theta_{t}}(x),w_y))^2 + \\ \beta \cdot \sum\nolimits_{c\neq y}(\max\{0, v- d(f_{\theta_{t}}(x),w_c)\}))^2,
    \label{eqn:loss_function}
\end{multline}
where the first term defines the positive part ($l_{pos}$) and the second term defines the negative part ($l_{neg}$) of the loss function. $d$ represents the distance function between the two input quantities. $\alpha$ and $\beta$ are hyper-parameters denoting the weight of each term. $v$ denotes the margin and $c$ and $y$ are the identity labels and the ground-truth identity label respectively. It essentially tries to accomplish the following:

\begin{itemize}
    \item Maximize the similarity between the instance embedding $f_{\theta}(x)$ and the positive class embedding $w_{y}$ where the instance $x$ belongs to the identity $y$.
    \item Minimize the similarity between the instance embedding $f_{\theta}(x)$ and the negative class embeddings $w_c$ for all identities $c \neq y$, where the instance $x$ belongs to the identity $y$.
\end{itemize}

At the face verification stage, we use the face feature extractor $f_{\theta}$ to extract the $d$-dimensional face representation. A distance metric (e.g., cosine similarity) between the two face features defines the similarity between the two face images. If the score is above a certain threshold $\tau$, we say that the two faces belong to the same identity; otherwise we decide that the two face images do not belong to the same identity. 

In the FL setup, we consider the realistic scenario where each of the client nodes actually correspond to individual user devices such as mobile phones which contain face images of only the owner of the device, rather than a central face dataset with multiple identities. Let us assume that there are $C$ mobile devices or client nodes. The server contains a face recognition system $f_{\theta_{0}}$ trained on a publicly available dataset. The $i^{th}$ client node contains $n_{i}$ instances from one identity (the owner of the device). Therefore, the $i^{th}$ client has access to the dataset $S^{i}$, where $S^{i} = {(x_{1}^{i},y^{i}),\ldots,(x_{n_{i}}^{i},y^{i})}$. Our objective is to enhance the performance of the face recognition system available on the server by collaboratively learning from the data on each of the clients without sharing training face images outside of the device.

\subsection{Federated Learning}
\label{section:FL}
FedAvg~\cite{fedavg} is one of the most widely adopted FL algorithm for training deep models. In FedAvg~\cite{fedavg}, a server facilitates the collaborative training of the model as follows: 

\begin{enumerate}
    \item In the $t^{th}$ communication round, the server sends the global model parameters $\theta_{t}$ and the class embedding matrix $W_{t}$ to all the client nodes.
    
    \item The $i^{th}$ client updates the global model at iteration $t$ based on the local data $S^{i}$ and the loss function 
    \begin{equation}
        L(f_{\theta_{t}};S^{i}) = \frac{1}{n_i}\sum\limits_{j \in [n_i]}l(f_{\theta_{t}}(x_{j}^{i}),y_{j}^{i})
    \end{equation}
    as follows
    \begin{equation}
        \theta_{t}^{i} = \theta_{t} - \eta \cdot \nabla_{\theta_{t}} L(f_{\theta_{t}};S^{i})
    \end{equation}
    \begin{equation}
        W_{t}^{i} = W_{t} - \eta \cdot \nabla_{W_{t}} L(f_{\theta_{t}};S^{i})
    \end{equation}
    
    \item The server receives the updated model parameters ($\theta_{t}^{i}$,$W_{t}^{i}$) from all the participating client nodes and updates the global model parameters by taking a weighted average of them as follows:
    \begin{align}
    \theta_{t+1} = \frac{1}{n}\sum_{i \in [C]}n_{i} \cdot \theta_{t}^{i}\ ; W_{t+1} = \frac{1}{n}\sum_{i \in [C]}n_{i} \cdot W_{t}^{i},
    \end{align}
where $n_{i}$ is the no. of training samples on the $i^{th}$ client and $n$ is the total no. of training samples across all client nodes. 
    
    \item Finally, the updated global parameters are transmitted to each of the clients and steps $2-4$ are repeated until convergence.
\end{enumerate}

Our objective is to enhance the face recognition performance of a pre-trained AFR model at the server by utilizing the face images available \emph{only} on end user devices (clients) such as mobile phones under the extreme case where each client has face images of only one identity. Conventional FL algorithms cannot be directly applied to such an instance because of two major reasons: 

\begin{itemize}
    \item In the $t^{th}$ communication round, the server cannot transmit the entire classification matrix $W_t$ to all users since the class embeddings contain sensitive information that can be potentially used to identify the user. Therefore, for the $i^{th}$ user, the server only communicates the parameters of the feature extractor $\theta_t$ and the class embedding $w_t^{i}$ associated with that user.
    
    \item The loss function (Eqn. ~\ref{eqn:loss_function}) encourages the instance embedding to be similar to the positive class embedding ($l_{pos}$) while at the same time minimizing the similarity between the instance embedding and the negative class embeddings ($l_{neg}$). In the current setting, the $i^{th}$ user only has access to face images of the $i^{th}$ identity and its positive class embedding $w_{i}$. Therefore, it cannot compute $l_{neg}$ while updating the model parameters on its local data as it does not have access to the negative class embeddings.
\end{itemize}

Optimizing only the positive part of the loss function ($l_{pos}$) would lead to a trivial solution where all the face images and the class embeddings collapse into a single point in the feature space resulting in sub-optimal performance. 

\subsection{Proposed Method}
\label{section:fedface}

To tackle the aforementioned challenges, an extra level of optimization at the server level is needed to ensure that the class embeddings are well separated in the feature space. To achieve this goal, we leverage the Spreadout regularizer~\cite{yu2020federated} to make sure that the embeddings of different identities are evenly distributed in the feature space. 

The proposed \emph{FedFace} framework is summarized in Algorithm~\ref{alg:fedface}. The server contains a pre-trained face extractor parametrized by $\theta_{0}$ trained on a publicly available dataset or a dataset readily available at the server. This is a reasonable assumption to make because there are celebrity face datasets such as Celeb-A~\cite{liu2015faceattributes} etc. available in the public domain, which can be used by the server for pre-training the system. The $i^{th}$ client/edge device contains a local face dataset $S_{i}$ with $n_{i}$ face images each of which belongs to the identity $i$. Therefore, in total we have $C$ clients and $C$ identities (each client holding multiple face images of one identity). \emph{FedFace} aims to enhance the performance of the pre-trained face feature extractor by collaboratively learning on the additional face images available on each of the clients. 

\begin{algorithm}[!t]
    \SetAlgoLined
    \SetKwInOut{Input}{Input}
    \SetKwInOut{Initialization}{Initialization}
    \SetKwInOut{Output}{Output}
    \Input{A pre-trained face feature extractor $f_{\theta_0}$ at the server and the local dataset $S^{i}$ consisting of $n_i$ face images belonging to the same identity $i$ at the $i^{th}$ client. In total, we have $C$ clients and $C$ identities with each client having multiple face images of one identity}
    \Initialization{T is the total number of communication rounds, $\eta$ is the learning rate at each client and $\lambda$ is the weight multiplier for the spreadout loss}
    The server initializes a global face feature extractor with the parameter $\theta_0$ \;
    The $i^{th}$ client initializes the class embedding $w_{0}^i$ with the mean feature as follows \break
    $w_{0}^{i} = \frac{1}{n_i}\sum\limits_{j=1}^{n_i}f_{\theta_{0}}(x_{j}^{i})$ \break
    $l_2$-normalize the class embeddings\;
    \For{$t = 1,\ldots,T$}{
        The server communicates the current global parameters $\theta_t$ and $w_t^i$ to the $i^{th}$ client (except for the first round where $w_0^i$ is initialized by the client itself)\;
        \For{each of the clients $i = 1, 2, \ldots, C$}{
        The client updates the model parameters based on the local data $S^i$\break
        $(\theta_{t+1}^i, w_{t+1}^{i}) \leftarrow (\theta_{t}^i, w_{t}^{i}) - \eta \nabla_{(\theta_{t}^i, w_{t}^{i})} L_{pos}(S^i)$\break
        where $L_{pos}(S^i) = \frac{1}{n_i}\sum\limits_{j=1}^{n_i}l_{pos}(f_{\theta_{t}^{i}}(x),i)$\;
        The $i^{th}$ client sends the updated parameters $(\theta_{t+1}^i, w_{t+1}^{i})$ back to the server
        }
        Server aggregates the updated parameters from all the clients\break
        $\theta_{t+1} \leftarrow \frac{1}{n}\sum\limits_{i \in [C]} n_{i} \cdot \theta_{t+1}^{i}$ \break
        where n represents the total number of samples on all the clients \break
        $\hat{W}_{t+1} = [w_{t+1}^1,\ldots,w_{t+1}^C]$\;
        Finally, the server employs the spreadout regularizer to separate the class embeddings from each other \break
        $W_{t+1} \leftarrow \hat{W_{t+1}} - \lambda \nabla_{\hat{W_{t+1}}} reg_{sp}(\hat{W_{t+1}})$
    }
    \Output{Output $\theta_{T}$}
    
    \caption{FedFace training procedure.}
    \label{alg:fedface}
\end{algorithm}

The server initializes a face feature extractor with the parameters $\theta_0$ and communicates this global model to each of the clients. In the first communication round, each of the clients need to initialize the class embedding $w_{0}^i$ for themselves which will then be locally updated using their local data along with the parameters $\theta_{t}$. Generally, the clients randomly initialize the class embedding for themselves from a Gaussian distribution. However, the class embeddings may lie outside the feature space of the pre-trained face extractor and thus, forcing the instance embeddings to be similar to the class embeddings can inhibit the already leaned embedding space by the pre-trained feature extractor and lead to inferior performance on general face recognition tasks. To alleviate this issue, \emph{FedFace} utilizes the mean of the embeddings of all the instances available at client $i$ extracted using the pre-trained feature extractor $f_{\theta_{0}}$ as the initialization of the class embeddings. Formally, if client $i$ has a dataset $S^{i} = {(x_{1}^{i}, y^{i}),\ldots,(x_{n_{i}}^{i}, y^{i})}$, the initialized class embedding for client $i$ can be denoted as :
\vspace{-0.1\baselineskip}
\begin{equation}
    w_{0}^{i} = \frac{1}{n_i}\sum\nolimits_{j=1}^{n_i}f_{\theta_{0}}(x_{j}^{i}).
\end{equation}

Since the mean feature is a linear combination of finite number of embeddings in $f_{\theta_{0}}$'s feature space, it also lies in the same feature space and thus provides a better initialization point for the class embeddings than the randomly initialized ones. We normalize the class embeddings to constrain them to lie in a $d$-dimensional hypersphere,~\ie, $||w_t^i||_2^2 = 1$.

In the $t^{th}$ communication round, the $i^{th}$ client then updates the parameters $\theta_t$ and $w_t^i$ by optimizing the positive loss function $l_{pos}$ using the local data $S^{i}$. \emph{FedFace} uses the squared hinge loss with cosine distance to define $l_{pos}$ at each of the individual clients
\vspace{-0.1\baselineskip}
\begin{equation}
    l_{pos}(f_{\theta_{t}}(x),i) = \max{(0, m - (w_{t}^{i})^{T}f_{\theta_{t}}(x))^2},
\end{equation}
where $m$ denotes the margin hyper-parameter and $w_i$ represents the class embedding for the $i^{th}$ client.

The updated values of the parameters $\theta$ and $w_i$ are communicated back to the server by all the clients. The server then aggregates the values of $\theta$ from each of the individual clients by taking a weighted average of them similar to the FedAvg algorithm. Since each of the clients only transmits the updated value of its own class embedding $w_t^i$, the server concatenates the class embeddings from all the clients to define the class embedding matrix $W_t$.
\vspace{-0.1\baselineskip}
\begin{equation}
    W_t = [w_t^1, w_t^2, \ldots, w_t^C]^T.
\end{equation}

Finally, to ensure that the class embeddings are well separated in the embedding space and do not collapse into a single point, we use the spreadout regularizer inspired by the FedAwS algorithm. After aggregating the class embeddings from all the clients in each communication round, the server optimizes the spreadout regularizer which is defined as follows:
\vspace{-0.1\baselineskip}
\begin{equation}
    {reg_{sp}(W_t)} = \sum\limits_{c\in[C]}\sum\limits_{\hat{c}\neq c}(\max\{0, v - d(w_t^c, w_t^{\hat{c}})\})^2,
\end{equation}
where $v$ represents the margin.

\begin{table*}[!t]
\footnotesize
\renewcommand{\arraystretch}{1.2}
\caption{Face Verification performance of \emph{FedFace} on standard face recognition benchmarks LFW~\cite{LFWTech}, IJB-A~\cite{klare2015pushing} and IJB-C~\cite{maze2018iarpa}.} 

% We use CosFace~\cite{wang2018cosface} ($64$-layer) as our feature extractor.}
\centering
\begin{tabularx}{\textwidth}{l|Y|Y|Y|Y}
\noalign{\hrule height 1.5pt}
              \multirow{2}{*}{\textbf{Method}} & \textbf{Training Data} & \textbf{LFW~\cite{LFWTech}} & \textbf{IJB-A~\cite{klare2015pushing}} & \textbf{IJB-C~\cite{maze2018iarpa}}\\ \cline{3-5}
             &  & \textbf{LFW Accuracy(\%)} & \textbf{TAR @ 0.1\% FAR} & \textbf{TAR @ 0.1\% FAR}\\ \hline
\noalign{\hrule height 1.0pt}
                Baseline & Centrally aggregated & $99.15\%$ & $81.43\%$ & $84.78\%$ \\
                Fine-tuning baseline in a non-federated manner & Centrally aggregated &  $99.32\%$ & $84.18\%$ & $88.76\%$\\
                Randomly Initialized class embeddings & Distributed & $94.61\%$ & $70.13\%$ & $69.30\%$ \\
                FedAvg~\cite{fedavg} & Distributed & $68.88\%$ & $19.75\%$ & $14.04\%$\\
                \noalign{\hrule height 1.0pt}
                \textbf{Proposed FedFace} & \textbf{Distributed} & \boldmath{$99.28\%$} &  \boldmath{$83.79\%$} & \boldmath{$88.21\%$} \\
\noalign{\hrule height 1.5pt}
\end{tabularx}
\label{tab:fgnet}
\vspace{-0.1in}
\end{table*}

The above steps are repeated for $T$ communication rounds at the end of which the global model parameter $\theta_T$ for the face feature extractor are returned. See Algorithm~\ref{alg:fedface} for the complete procedure.

\section{Experimental Results}

\subsection{Experimental Settings}
\noindent \textbf{Datasets : } We train \emph{FedFace} using face images in CASIA-WebFace~\cite{yi2014learning} dataset. CASIA-WebFace comprises of $494,414$ images from $10,575$\footnote{We remove 84 subjects in CASIA-WebFace that overlap with LFW} different subjects. We randomly choose $10,000$ subjects from the CASIA-WebFace dataset as our training set. To test the performance of \emph{FedFace} on Face Verification task, we evaluate on three different face benchmarks, namely LFW~\cite{LFWTech}, IJB-A~\cite{klare2015pushing} and IJB-C~\cite{maze2018iarpa}. LFW~\cite{LFWTech} contains $13,233$
face images of $5,749$ subjects. We evaluate \emph{FedFace} on LFW under the standard LFW protocol. IJB-A~\cite{klare2015pushing} utilizes a template-based benchmark, containing $25,813$ faces images of $500$ subjects. Each template includes a set of still photos or video frames. IJB-C~\cite{maze2018iarpa} is an extension of IJB-A with $140,740$ faces images of $3,531$ subjects. Note that IJB-C is not a more challenging dataset than IJB-A and their performances are comparable in the literature.\\

\begin{figure}[!t]
\captionsetup{font=footnotesize}
\scriptsize
% \settoheight{\tempdima}{\includegraphics[width=0.3\linewidth]{example-image-a}}%
\centering
\begin{subfigure}[]{\linewidth}
\centering
\begin{tabular}{@{}c@{ }c@{ }c@{ }c@{ }c@{ }}
\includegraphics[width=0.24\textwidth]{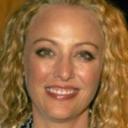} &
\includegraphics[width=0.24\textwidth]{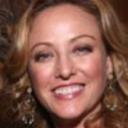} &
\includegraphics[width=0.24\textwidth]{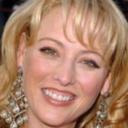} &
\includegraphics[width=0.24\textwidth]{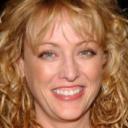}
\end{tabular}
\vspace{-1\baselineskip}
\caption{CASIA-WebFace~\cite{yi2014learning}}
\vspace{5pt}
\end{subfigure}
\begin{subfigure}[]{\linewidth}
\centering
\begin{tabular}{@{}c@{ }c@{ }c@{ }c@{ }c@{ }}
\includegraphics[width=0.24\textwidth]{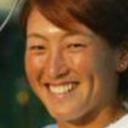} &
\includegraphics[width=0.24\textwidth]{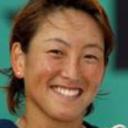} &
\includegraphics[width=0.24\textwidth]{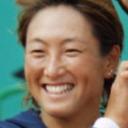} &
\includegraphics[width=0.24\textwidth]{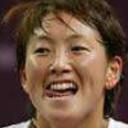}
\end{tabular}
\vspace{-1\baselineskip}
\caption{LFW~\cite{LFWTech}}
\vspace{5pt}
\end{subfigure}
\begin{subfigure}[]{\linewidth}
\centering
\begin{tabular}{@{}c@{ }c@{ }c@{ }c@{ }c@{ }}
\includegraphics[width=0.24\textwidth]{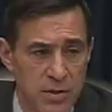} &
\includegraphics[width=0.24\textwidth]{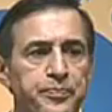} &
\includegraphics[width=0.24\textwidth]{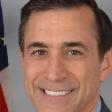} &
\includegraphics[width=0.24\textwidth]{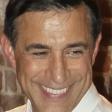}
\end{tabular}
\vspace{-1\baselineskip}
\caption{IJB-A~\cite{klare2015pushing}}
\vspace{5pt}
\end{subfigure}
\begin{subfigure}[]{\linewidth}
\centering
\begin{tabular}{@{}c@{ }c@{ }c@{ }c@{ }c@{ }}
\includegraphics[width=0.24\textwidth]{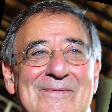} &
\includegraphics[width=0.24\textwidth]{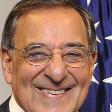} &
\includegraphics[width=0.24\textwidth]{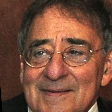} &
\includegraphics[width=0.24\textwidth]{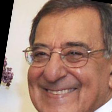}
\end{tabular}
\vspace{-1\baselineskip}
\caption{IJB-C~\cite{maze2018iarpa}}
\vspace{5pt}
\end{subfigure}
\caption{Examples of face images from (a) CASIA-WebFace~\cite{yi2014learning}, (b) LFW~\cite{LFWTech}, (c) IJB-A~\cite{klare2015pushing} and (d) IJB-C~\cite{maze2018iarpa} datasets. Each row consists of face images of one identity.}%
\label{fig:Dataset_examples}
\end{figure}

\noindent \textbf{Implementation :} For the face feature extractor, we use a publicly available face matcher, CosFace~\cite{wang2018cosface} ($64$ layer), which outputs a $512$-dimensional feature vector. For pre-training the global face recognition model on the server, we use an angular margin softmax loss with the scale hyper-parameter value of $30$ and the margin hyper-parameter value of $10$ similar to CosFace. For training in the FL setup at each individual client, we use a squared hinge loss with cosine distance with a margin of $0.9$ as the positive loss function. We normalize both the instance embeddings and the class embeddings to make their norm as $1$. We train the model for $200$ communication rounds. We choose a learning rate of $0.001$ to update the model locally on each of the clients. We use a small learning rate to make sure that training in the collaborative setting does not deviate much from the original embedding space. We use a sufficiently large weight for the spreadout regularizer $\lambda=10$) to ensure that the class embeddings are well separated.

\subsection{Effect of number of clients}

\begin{figure}[!t]
    \centering
    \captionsetup{font=footnotesize}
    \includegraphics[width=\linewidth]{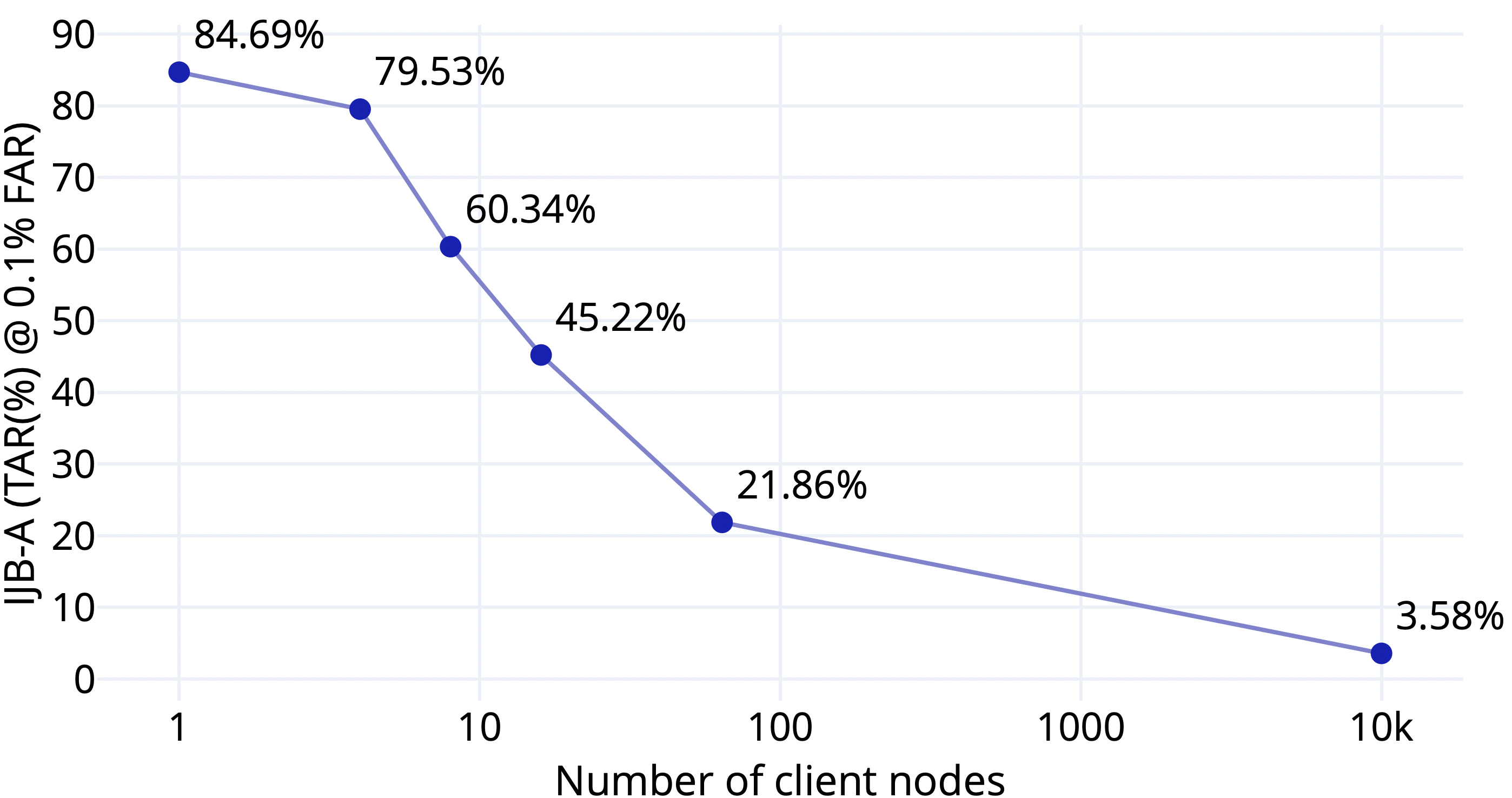}
    \caption{Effect of the number of clients on the FedAvg~\cite{fedavg} algorithm. We divide the $10,000$ subjects in CASIA-WebFace~\cite{yi2014learning} equally into different client nodes for training. One client node denotes the conventional (non-federated) way of training AFR systems while the $10k$ clients represents the problem we are tackling with face images of one identity per client. We evaluate on IJB-A~\cite{klare2015pushing}. Note that the x-axis is in log scale.}
    \label{fig:clients_effect}
\end{figure}

To evaluate the effect of the number of clients participating in the collaborative learning on the performance of the conventional FedAvg~\cite{fedavg} algorithm, we divide our training set ($10,000$ subjects of CASIA-WebFace~\cite{yi2014learning}) into a number of clients with equal number of subjects available for training at each client. One client scenario represents the traditional way of learning face recognition models where all the data is available centrally. We test on $4$ ($2500$ subjects on each client), $8$ ($1,250$ subjects on each client), $16$ ($625$ subjects on each client), $64$ ($156$ subjects on each client) and finally $10,000$ ($1$ subject on each client) client nodes. Fig.~\ref{fig:clients_effect} represents the performance of FedAvg~\cite{fedavg} on IJB-A~\cite{klare2015pushing} dataset under these settings. Note that since FedAvg uses the AM-Softmax loss function for training on the clients without the spreadout regularizer, it does not enforce the class embeddings to be well separated. Therefore, in the $10,000$ clients case, FedAvg leads to a trivial solution with the feature vectors for all the face images clustering around a single point deteriorating the performance significantly. 
% \emph{FedFace}, on the other hand, explicitly enforces the class embeddings to be well separated by employing the spreadout regularizer and thus improves the performance of the system.

\begin{figure}
    \centering
    \includegraphics[width=0.23\linewidth]{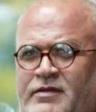}
    \includegraphics[width=0.23\linewidth]{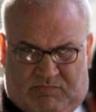}\hfill
    \includegraphics[width=0.23\linewidth]{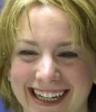}
    \includegraphics[width=0.23\linewidth]{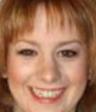}\\
   (a) \hspace{10em}(b)
    \caption{Examples of two genuine pairs of face images from LFW dataset, (a) Saeb Erekat and (b) Sarah Hughes which failed to match using the pre-trained face feature extractor. After training on the additional $1,000$ subjects using \emph{FedFace}, the two genuine pairs were correctly matched.}
    \label{fig:verification_example}
\end{figure}

\subsection{Performance of FedFace on Standard Face benchmarks}

To evaluate the performance of \emph{FedFace}, we divide our training set into two parts. $9,000$ of the $10,000$ subjects in the training set are used to pre-train a face recognition system on the server. This set simulates publicly available face datasets or face images available on the server for pre-training. The remaining $1,000$ subjects are distributed into $1,000$ clients with each client having access to all the face images of one subject. We evaluate \emph{FedFace} in this setting on three different face benchmarks namely LFW~\cite{LFWTech}, IJB-A~\cite{klare2015pushing} and IJB-C~\cite{maze2018iarpa}. We compare the proposed \emph{FedFace} (Section \ref{section:fedface}) with the following methods:
% \vspace{0.2em}

\noindent \textbf{Baseline:} Pre-training CosFace with face images of the $9,000$ subjects from CASIA-WebFace. This is the face recognition model that the server has access to initially.
% \vspace{0.2em}

\noindent \textbf{Fine-tuning baseline in a non-federated manner:} Centrally aggregating the face images of the $1,000$ clients and using them to boost the performance of the pre-trained face recognition model in a non-federated manner. This is the traditional way of training DNN based face recognition models but suffer from serious privacy concerns since the face images are shared to the server.
% \vspace{0.2em}

\noindent \textbf{Randomly initialized class embeddings:} As explained in Section~\ref{section:FL}, training with only the positive loss function ($l_{pos}$) can lead to a trivial solution where all the class embeddings and face feature vectors collapse into a single point in the embedding space. One way to counter this is to randomly initialize the class embeddings and fixing them during training. This prevents the class embeddings from collapsing into single point but does not explicitly force them to be well separated.
since the face images are shared to the server.
% \vspace{0.2em}

\noindent \textbf{FedAvg:} The conventional FedAvg~\cite{fedavg} algorithm where the AFR system is trained in the FL setup but with only the positive part of the loss function ($l_{pos}$).

Table~\ref{tab:fgnet} shows the comparison of \emph{FedFace} with the above methods. \emph{FedFace} is able to boost the performance of the pre-trained face recognition system by collaboratively learning from the additional face images available on each of the individual clients. It effectively bridged the gap between the non-federated setup and the federated setup of learning while ensuring that the face images available on each of the clients are kept private. Fig.~\ref{fig:verification_example} shows two genuine pairs of face images from the LFW dataset which failed to match using the pre-trained face feature extractor but after training on the additional $1,000$ subjects using FedFace, they were matched correctly.

\subsection{Ablation Study}

We ablate over the mean feature initialization scheme and the spreadout regularizer to study their individual effect on the performance of \emph{FedFace}. The results of the ablation study are summarized in Table~\ref{tab:ablation}. Note that although, the mean feature initialization does not help significantly in boosting the final performance, it helps in faster convergence (200 communication rounds compared to 450 communication rounds with random initialization) and is therefore, extremely crucial for such a collaborative learning framework where the communication latency between the clients and the server is usually a critical bottleneck.

\begin{table}[h]
\scriptsize
\caption{Ablation Study}
\centering
\begin{tabularx}{\linewidth}{Y|Y|Y|Y}
\noalign{\hrule height 1.5pt}
     \multirow{2}{*}{} & \textbf{LFW~\cite{LFWTech}} & \textbf{IJB-A~\cite{klare2015pushing}} & \textbf{IJB-C~\cite{maze2018iarpa}}\\
             & \textbf{LFW Accuracy} & \textbf{TAR @ 0.1\% FAR} & \textbf{TAR @ 0.1\% FAR}\\ \hline
    \textbf{w/o mean feature initialization} & $99.28\%$ & $83.73\%$ & $88.12\%$ \\\hline
    \textbf{w/o Spreadout Regularizer} & $75.90\%$ & $34.16\%$ & $25.40\%$\\
\noalign{\hrule height 1.5pt}
\end{tabularx}
\vspace{-1.5em}
\label{tab:ablation}
\end{table}

\section{Conclusion}

In this paper, we tackle the problem of learning a collaborative face recognition system from face images available on a collection of mobile devices in a privacy aware manner. We propose a federated learning framework called \emph{FedFace} to enhance the performance of a pre-trained face recognition system, CosFace, under the extreme case where each mobile device has face images of only one identity. We evaluate our approach on three standard face recognition benchmarks namely LFW, IJB-A and IJB-C. The proposed approach is able to enhance the performance of CosFace by effectively utilizing the additional data available on mobile devices without the face images ever leaving the device. In the future, we aim to extend our work to adapt a pre-trained AFR system to new domains such as selfie images using face images available on mobile devices.

\section*{Acknowledgement}
\noindent This research was supported in part by ONR grant N00014-20-1-2382.

{\footnotesize
\bibliographystyle{ieee}
\bibliography{egbib}
}

\end{document}

%%%%%%%%% TITLE
\title{FedFace: Collaborative Learning of Face Recognition Model\\
(Supplementary Material)}

\author{Divyansh Aggarwal\\
Michigan State University\\
East Lansing, MI, USA\\
{\tt\small aggarw49@msu.edu}
\and
Jiayu Zhou\\
Michigan State University\\
East Lansing, MI, USA\\
{\tt\small jiayuz@msu.edu}

\and
Anil K. Jain\\
Michigan State University\\
East Lansing, MI, USA\\
{\tt\small jain@msu.edu}
}

\maketitle
\thispagestyle{empty}

In this supplementary material, we include more implementation  details of \emph{FedFace}.

\section{Implementation Details}
All models are implemented using Tensorflow r1.14.0. A single NVIDIA GeForce RTX 2080 Ti GPU is used for training and testing.

\subsection{Data Preprocessing}

All face images are passed through MTCNN face detector~\cite{mtcnn} to detect five landmarks (two eyes, nose, and two mouth corners). Via similarity transformation, the face images are aligned. After transformation, the images are resized to $112\times 96$.

\subsection{Face Feature Extractor}

In all our experiments, for our face feature extractor $f_{\theta}$, we use a $64$-layer residual network similar to CosFace~\cite{wang2018cosface} which outputs a $512$-dimensional feature vector. Table~\ref{tab:architecture} shows the architecture of our face feature extractor. Conv1, Conv2, Conv3 and Conv4 denote $4$ convolution units that contain multiple convolution layers and residual units. Residual units are enclosed in double-column brackets. Eg., $[3\times3,64]$ denotes $4$ cascaded convolutional layers with $64$ filters of size $3\times3$, and $S2$ denotes stride $2$. $FC1$ is the fully connected layer.

\begin{table}[!t]
\footnotesize
\renewcommand{\arraystretch}{1.5}
\caption{Architecture of the $64$-layer residual network we use for our face feature extractor}
\centering
\begin{tabularx}{\linewidth}{|X|X|}
\noalign{\hrule height 1.5pt}
    \textbf{Layer} & \\
    \noalign{\hrule height 1.5pt}
    \multirow{3}{*}{\textbf{Conv1}} & $[3\times3, 64] \times 1, S2$\\
    & $\begin{bmatrix}
        3\times3, 64\\
        3\times3, 64
    \end{bmatrix}$ $\times 3$\\
    \noalign{\hrule height 1pt}
    \multirow{3}{*}{\textbf{Conv2}} & $[3\times3, 128] \times 1, S2$\\
    & $\begin{bmatrix}
        3\times3, 128\\
        3\times3, 128
    \end{bmatrix}$ $\times 8$\\
    \noalign{\hrule height 1pt}
    \multirow{3}{*}{\textbf{Conv3}} & $[3\times3, 256] \times 1, S2$\\
    & $\begin{bmatrix}
        3\times3, 256\\
        3\times3, 256
    \end{bmatrix}$ $\times 16$\\
    \noalign{\hrule height 1pt}
    \multirow{3}{*}{\textbf{Conv4}} & $[3\times3, 512] \times 1, S2$\\
    & $\begin{bmatrix}
        3\times3, 512\\
        3\times3, 512
    \end{bmatrix}$ $\times 3$\\
    \noalign{\hrule height 1pt}
    \textbf{FC1} & 512\\
\noalign{\hrule height 1.5pt}
\end{tabularx}
\label{tab:architecture}
\end{table}

\subsection{Training Details}

We use an angular margin softmax loss (AM-Softmax) with the scale hyper-parameter value of $30$ and the margin hyper-parameter value of $10$ similar to CosFace~\cite{wang2018cosface} for training our face feature extractor in the non-federated setup. For training in the FL setup at each individual client, we use a squared hinge loss with cosine distance with a margin of $0.9$ as the positive loss function. We normalize both the instance embeddings and the class embeddings to make their norm as $1$. We train the model for $200$ communication rounds with a SGD optimizer with momentum $0.9$. We choose a learning rate of $0.001$ to update the model locally on each of the clients. Empirically, we set the weight of the spreadout regularizer, $\lambda=10$.

%%%%%%%%% ABSTRACT

% \twocolumn[{%
% \renewcommand\twocolumn[1][]{#1}%
% \maketitle
%  \thispagestyle{empty}
% \begin{center}
%     \includegraphics[width=\linewidth]{images/main_figure.jpg}
% \end{center}
%     \captionof{Figure 1: }{An overview of the federated setup for training face recognition model. We tackle the scenario where each client is a mobile device with face images pertaining to only the user/owner of the device rather than data centers or organizations holding face datasets with multiple identities. The server sends the current global model to each of the participating client nodes who update the model based on their local training data. These local updates are transferred back to the server where they are aggregated to generate a global update. Through several rounds of communication between the servers and the client, a collaborative face recognition system can be obtained in a privacy preserving manner.}
%     \label{fig:Overview}
%     \vspace{1em}

% }]

{\footnotesize
\bibliographystyle{ieee}
\bibliography{egbib}
}